\documentclass[preprint]{elsarticle}

\usepackage{amssymb}
\usepackage{amsmath}
\usepackage{lineno}
\usepackage{dsfont}
\usepackage{algorithmic}
\usepackage{algorithm}
\usepackage{array}
\usepackage{hyperref}
\usepackage{xurl}
\usepackage{textcomp}
\usepackage{stfloats}
\usepackage{hyperref}       % hyperlinks
\usepackage{url}            % simple URL typesetting
\usepackage{booktabs}       % professional-quality tables    % blackboard math symbols
\usepackage{nicefrac}       % compact symbols for 1/2, etc.
\usepackage{microtype}      % microtypography
\usepackage{xcolor}         % colors
\usepackage{graphicx}
\usepackage{cite}
\usepackage{multirow}
\usepackage{xcolor}
\usepackage{threeparttable}
\usepackage[capitalise]{cleveref}
\usepackage{tikz}
\usepackage{siunitx}

\journal{Vestnik of Saint Petersburg University. Applied Mathematics. Computer Science. Control Processes}

\makeatletter
\def\ps@pprintTitle{%
  \let\@oddhead\@empty
  \let\@evenhead\@empty
  \def\@oddfoot{\reset@font\hfil\parbox{1\textwidth}{\footnotesize\itshape Preprint submitted to \@journal}\hfil}%
  \let\@evenfoot\@oddfoot}
\makeatother

\begin{document}

\begin{frontmatter}



\title{\LARGE{Neural Operators for Mathematical Modeling of Transient Fluid Flow in Subsurface Reservoir Systems}}

\author[1,2]{Daniil D. Sirota} %\corref{cor1}
\affiliation[1]{organization={PJSJ Gazprom}, city={Saint Petersburg}, country={Russian Federation}}
\affiliation[2]{organization={St. Petersburg State University}, city={Saint Petersburg}, country={Russian Federation}}

\author[1]{Sergey A. Khan}

\author[1]{Sergey L. Kostikov}

\author[1]{Kirill A. Butov}


% \author{} %% Author name

% \affiliation{organization={},%Department and Organization

%% Abstract
\begin{abstract}

This paper presents a method for modeling transient fluid flow in subsurface reservoir systems based on the developed neural operator architecture (TFNO-opt). Reservoir systems are complex dynamic objects with distributed parameters described by systems of partial differential equations (PDEs).
Traditional numerical methods for modeling such systems, despite their high accuracy, are characterized by significant time costs for performing calculations, which limits their applicability in control and decision support problems.
The proposed architecture (TFNO-opt) is based on Fourier neural operators, which allow approximating PDE solutions in infinite-dimensional functional spaces, providing invariance to discretization and the possibility of generalization to various implementations of equations. The developed modifications are aimed at increasing the accuracy and stability of the trained neural operator, which is especially important for control problems. These include adjustable internal time resolution of the integral Fourier operator, tensor decomposition of parameters in the spectral domain, use of the Sobolev norm in the error function, and separation of approximation errors and reconstruction of initial conditions for more accurate reproduction of physical processes. The effectiveness of the proposed improvements is confirmed by computational experiments.
The practical significance is confirmed by computational experiments using the example of the problem of hydrodynamic modeling of an underground gas storage (UGS), where the acceleration of calculations by six orders of magnitude was achieved, compared to traditional methods. This opens up new opportunities for the effective control of complex reservoir systems.

\end{abstract}


% \begin{highlights}
% \end{highlights}

% %% Keywords
\begin{keyword}
% %% keywords here, in the form: keyword \sep keyword

% %% PACS codes here, in the form: \PACS code \sep code

% %% MSC codes here, in the form: \MSC code \sep code
% %% or \MSC[2008] code \sep code (2000 is the default)
mathematical modeling \sep deep learning \sep neural networks \sep neural operators \sep hydrodynamic modeling \sep subsurface reservoir \sep fluid flow in porous medium
\end{keyword}

\end{frontmatter}



%\noindent{\it Keywords}: Lithium, POMDP, Decision-Making, %Energy Transition, Critical Minerals, Supply Chain
% \maketitle
% \ioptwocol
%

\section{Introduction}
Reservoir systems are ensembles of geological bodies (formations) composed of a porous medium saturated with fluids (oil, gas, water) and subject to physical fields. These systems are characterized by interactions among formations, wells, fluids, and the surrounding geological medium, forming a dynamically coupled system~[1].

Traditional approaches to the mathematical modeling of reservoir systems can be divided into analytical (material balance) methods and numerical hydrodynamic models (reservoir simulators). 
Material balance methods, based on simplified analytical dependencies, do not account for the geological heterogeneity of reservoirs and the spatio-temporal dynamics of flow processes in porous media, which significantly reduces their accuracy. Numerical models, which rely on solving systems of partial differential equations (PDEs), provide a detailed reproduction of the dynamics of these flow processes. However, achieving high accuracy is associated with substantial computational costs: the solution time for typical problems can be several hours, even when using high-performance computing systems.

These characteristics significantly narrow the practical applicability of numerical models where multiple simulations are required, which is critical for decision-making and optimal control problems.

Deep learning methods, particularly neural networks, can be considered an alternative to traditional approaches for the numerical modeling of physical systems. However, most research in this area is devoted to learning mappings in finite-dimensional spaces~[2, 3], whereas dynamic processes in reservoir systems, described by PDEs, require the construction of mappings in infinite-dimensional function spaces.

According to the universal approximation theorems~[4, 5], fully-connected neural networks with sufficient complexity can approximate any continuous function on compact sets to a given accuracy. This result was extended in~[6, 7] to nonlinear operators mapping between infinite-dimensional function spaces, and in~[8], quantitative estimates of the approximation error's dependence on the problem's dimensionality and the number of model parameters were obtained. However, the theoretical possibility of approximating mappings in function spaces does not specify methods for their effective practical implementation. It is known that neural network architectures show significantly different performance depending on the class of problems: for example, fully-connected networks are considerably inferior to convolutional networks in image processing~[3].

The application of neural networks for approximating solutions to PDEs faces the "curse of dimensionality"~[9], especially in problems with complex geometry (e.g., heterogeneous reservoirs) or high-dimensional parameter spaces (in particular, the equations for flow in porous media)~[1]. As shown in~[10], the upper bound estimate for the generalization error of neural network models has an asymptotic behavior of \( \mathcal{O}(N^{-1/2}) \), where \(N\) is the size of the training set. To achieve an accuracy of 1\%, a training set size of \( N \sim 10^4 \) is required, which is often unattainable when using resource-intensive numerical simulations. These limitations highlight the relevance of developing specialized neural network architectures for such tasks. Currently, three main approaches are distinguished in the field of approximating PDEs with deep learning methods.

1. \textit{Traditional architectures} (CNN, RNN, GAN)~[11–13], trained on data from numerical simulations. These methods exhibit dependence on the geometry of the modeled domain, the discretization grid, and require significant amounts of training data for accurate approximation.

2. \textit{Physics-informed neural networks} (PINN)~[14, 15]. With these, the equations are embedded into the training process by minimizing the residuals of the PDEs and boundary conditions using automatic differentiation~[16]. However, PINNs only approximate a specific instance of the equation (e.g., a combination of parameters, initial, and boundary conditions), which does not provide a computational advantage over classical methods for solving PDEs in many applied problems.

3. \textit{Neural Operators} (NO)~[17--19]. This approach formalizes neural operators as mappings between infinite-dimensional function spaces, thereby ensuring discretization invariance due to universal approximation properties. After being trained on a limited dataset, they are applicable to different discretization grids and PDE realizations, including equations with variable coefficients and nonlinear terms.

The goal of this work is to develop an effective neural operator architecture aimed at overcoming the computational limitations of numerical hydrodynamic models of reservoir systems.

\textbf{\textit{Main contribution of the work.}} This work proposes a neural operator architecture (TFNO-opt) adapted for modeling transient flow in porous media in reservoir systems. The key features are:
\begin{itemize}
\setlength{\itemsep}{0pt}
\setlength{\parskip}{2pt}
\item The adjustable temporal resolution of the integral operator's kernel and the tensor decomposition of parameters enhance modeling accuracy while significantly reducing the number of parameters;

\item The use of a Sobolev norm and the separation of approximation and initial condition reconstruction errors ensure that the trained model accurately reproduces the physical nature of the processes.
\end{itemize}
\textbf{\textit{Practical significance.}} The developed model provides a speed-up in calculations of at least six orders of magnitude compared to a numerical simulator, as demonstrated on a model of an underground gas storage (UGS) facility.

\section{Methodology}

\subsection{Mathematical Model of the Flow Process}
Consider the process of transient gas flow in a porous medium, which in its general form~[20, 21] can be described by an equation for an unknown function $u=u(\mathbf{x}, t)$ (e.g., density or pressure), dependent on spatial coordinates $\mathbf{x} \in \Omega \subset \mathbb{R}^d$ ($d=1,2,3$) and time $t > 0$:
\begin{equation}
\label{eq:zero}
\partial_t u = \nabla \cdot \left( \nabla \Phi(u) \right) + f = \nabla \cdot \left( D(u)  \nabla u \right) + f.
\end{equation}
Here, 
\( \Phi(u) \) is a nonlinear increasing function related to the flux;
\( D(u) = \partial \Phi / \partial u\) is the diffusion coefficient;
\(f= f(\mathbf{x}, t) \) is the source/sink function.
Structurally, (\ref{eq:zero}) is a quasilinear parabolic equation.

We will characterize the mathematical model based on the equation for three-dimensional, transient, single-phase flow of a compressible fluid (natural gas) in a porous medium. This equation is derived by substituting the momentum conservation law (Darcy's law of flow) into the mass conservation law. Taking into account second-type boundary conditions (no-flow across the reservoir boundaries) and the initial pressure distribution, the mathematical model can be written as follows~[22, p. 68]:

\begin{equation}
\label{eq:one}
\left\{
\begin{aligned}
&\nabla \cdot \left( \frac{A \mathbf{k}}{\mu_g B_g} \nabla p \right) \Delta \textbf{x} = \frac{V_b \phi T_{\text{sc}}}{p_{\text{sc}} T_{\text{res}}} \frac{\partial}{\partial t}\left(\frac{p}{Z}\right) - q \ \text{in } \Omega \times (0, T], \\
&\left. \frac{\partial p}{\partial n} \right|_{\Gamma} = 0 \ \text{on } \Gamma, \\
&p(\cdot, 0)  = p_0 \ \text{at } t = 0,
\end{aligned}
\right.
\end{equation}
here,
\( p \) is the pressure,
\( q \) is the gas flow rate at standard conditions, \( \nabla = \left( \frac{\partial}{\partial x}, \frac{\partial}{\partial y}, \frac{\partial}{\partial z} \right) \) is the nabla operator,
\( \mathbf{k} = (k_x, k_y, k_z) \) is the permeability tensor,
\( A = (A_x, A_y, A_z) \) are the cross-sectional areas of flow along the respective axes,
\( \Delta\textbf{x} = (\Delta x, \Delta y, \Delta z) \) is the vector of control volume dimensions,
\( B_g = \frac{p_{\text{sc}}TZ}{T_{\text{sc}}p} \) is the gas formation volume factor,
\( Z \) is the compressibility factor (\textit{z}-factor) of a real gas,
\( \mu_g \) is the gas viscosity,
\( T_{\text{sc}} \) and \( p_{\text{sc}} \) are the temperature and pressure at standard conditions,
\( T_{\text{res}} \) is the reservoir temperature,
\( \phi \) is the porosity,
\( V_b \) is the control volume,
\( \Omega \subset \mathbb{R}^3 \) is a bounded open set (the solution domain),
\( \Gamma = \partial \Omega \) is the boundary of the solution domain, satisfying a Lipschitz condition,
\(n\) is the normal to the boundary,
\( T > 0 \) is a fixed time,
\( p(\cdot, 0)\) is the initial pressure distribution.

Equation (\ref{eq:one}) can only be solved numerically~[22, p. 68] and is formulated in a form adapted for solution by the finite volume method, where the parameters \( A \), \( \Delta \mathbf{x} \), and \( V_b \) correspond to the geometry of the control volume. This approach is standard in the field of reservoir system modeling.

Within the accepted assumptions, it is presumed that all coefficients in (\ref{eq:one}) consist of bounded, measurable, and positive definite functions. The pressure is bounded \((0<p_{\text{min}}<p<p_{\text{max}})\), the nonlinear parameters are monotonic Lipschitz functions of pressure, with \(\mu_g(p) > 0\), \(B_g(p) > 0\), and \(Z(p) > 0\) for all \(p\). The tensor \(\mathbf{k}\) satisfies the condition of uniform ellipticity, and the well flow rates are bounded \(( q_{\text{min}}<q<q_{\text{max}})\). The function \(p/z \) is strictly monotonically increasing with an increase in pressure~[22].

The existence and uniqueness of a weak solution to equation (\ref{eq:one}) in the function class \(p \in L^2(0, T ; H^1(\Omega )) \cap H^1(0, T; H^{-1}(\Omega)) \) for \(p_0 \in L^2(\Omega) \) and \(q \in L^2(0,T;L^2(\Omega))\) follow from the classical results of the theory of quasilinear parabolic equations. In particular, the well-posedness of the transient flow equation is justified in~[20, p. 693; 21, p. 102] under conditions of uniform ellipticity, Lipschitz continuity, and boundedness of the coefficients.

A common approach to modeling sources and sinks is their direct inclusion in the flow equation~[23, p. 67]. Then, the well model for \(q \in L^2(0,T; L^2{(\Omega}))\) can be formulated as follows \eqref{eq:well}:
\begin{equation}
\label{eq:well}
q(t) =  \sum_{m=1}^{M_{w}}\sum_{\nu=1}^{N_{w\nu}} q_m^{(\nu)}(t) \eta_\epsilon\left(\mathbf{x} - \mathbf{x}_m^{(\nu)}\right).
\end{equation}

In \eqref{eq:well}, \( \mathbf{x}_m^{(\nu)} \in \Omega \) are the coordinates of the well perforation zones,
\( q_m^{(\nu)}(t) \in L^2(0,T) \) is the source/sink intensity in the \( \nu \)-th perforated zone of the \( m \)-th well at time \( t \), \( \eta_\epsilon(\mathbf{x}) \in L^2(\Omega) \) is a smooth approximation of the delta function, \( \epsilon > 0 \), \( M_{w} \) is the number of wells, and \( N_{{w\nu}} \) is the number of well perforation zones. It is assumed that the wells do not intersect in space.

It should be noted that there are numerous models describing flow processes in porous media. The approaches and methods proposed in this work are universal in nature and do not depend on specific mathematical models of flow. This is because they are based on the results of numerical simulations and the universal approximation property of neural operators, rather than on the specifics of the models themselves. Consequently, these methods can be applied to both single-phase and multiphase flows, provided that appropriate clarifications and assumptions are made.

\subsection{Operator Learning}
Consider the problem of approximating the solution operator of a PDE by constructing and training a neural operator. We introduce a mapping \(\mathcal{G}\) that acts as the \textit{solution operator} of the PDE:

\begin{equation}
\label{eq:two}
\begin{aligned}
     \mathcal{G}: \mathcal{A}\left(D;\mathbb{R}^{d_a}\right) &\rightarrow \mathcal{U}\left(D;\mathbb{R}^{d_u}\right) ,\\
     a &\mapsto u := \mathcal{G}(a),
\end{aligned}
\end{equation}
here, \( D=\Omega \times [0, T] \subset \mathbb{R}^d \) is the domain of definition; \( \mathcal{A}\left(D;\mathbb{R}^{d_a}\right)\) and \( \mathcal{U}\left(D;\mathbb{R}^{d_u}\right) \) are Banach spaces of functions defined on \( D \) with values in \( \mathbb{R}^{d_a} \) and \( \mathbb{R}^{d_u} \) respectively; \( a \in \mathcal{A}\left(D;\mathbb{R}^{d_a}\right) \) is the input function, which can include initial conditions, boundary conditions, and parameters of the differential equation, i.e., \( a: D \mapsto \mathbb{R}^{d_a} \); and \( u \in \mathcal{U}\left(D;\mathbb{R}^{d_u}\right) \) is the output function, which is the solution to the PDE, i.e., \( u: D \mapsto \mathbb{R}^{d_u} \). The operator \(\mathcal{G}\) maps each input function \textit{a} to the solution of the differential equation \textit{u}. In the formulation (\ref{eq:two}), the problem of approximating the operator \(\mathcal{G}\) consists of constructing a mapping that reproduces the solutions of the PDE with a given accuracy.

To train \(\mathcal{G}\), a finite set of data pairs \(\{a_j, u_j\}_{j=1}^N\) is used, where \({a}_j \sim \mu\) is a sequence of independent and identically distributed random variables obtained from a probability measure \(\mu\) defined on \(\mathcal{A}\), and \({u}_j = \mathcal{G}({a}_j)\) is the corresponding PDE solution. The data can be obtained from actual observations or through numerical simulations.

We define a \textit{neural operator} as a parameterized mapping \(\mathcal{G}_\theta\) that approximates the operator \(\mathcal{G}\), where \(\theta \in \Theta\) is a set of trainable parameters. The optimization of parameters \(\theta\) is performed by minimizing a loss function \(L : \mathcal{U}(D; \mathbb{R}^{d_u}) \times \mathcal{U}(D; \mathbb{R}^{d_u}) \ \to \mathbb{R_+}\). Formally, the learning problem can be written as follows:

\begin{equation*}
\label{eq:loss_1}
\theta^* = \arg\min_{\theta \in \Theta} \mathbb{E}_{{a\sim\mu}} \left[ L\left(\mathcal{G}_\theta(a), u\right) \right],      
\end{equation*}
where \(\mathbb{E}\) is the mathematical expectation.

Thus, the problem is reduced to finding the optimal values of the parameters \(\theta^*\in\Theta\) that minimize the expected value of the loss function.

\section{Development of the Neural Operator Architecture}
The main objectives of this work include the development of an architecture and the training of a neural operator capable of effectively approximating the solution of the PDE system \eqref{eq:one}. The construction of a neural operator typically involves the following stages~[18, p. 9]:

1) Application of an operator $\mathcal{P}$ that maps the input data into a higher-dimensional hidden space;

2) Iterative application of an integral transform operator $\mathcal{L}$;

3) Projection into the physical space using an operator $\mathcal{Q}$.

Thus, the general structure of neural operators takes the form:
\begin{equation}\label{eq:four}  
\mathcal{G}_\theta(a) = \mathcal{Q} \circ \mathcal{L}_L \circ \cdots \circ \mathcal{L}_1 \circ \mathcal{P}(a).  
\end{equation}  
Here, $\mathcal{P}: \mathcal{A}(D; \mathbb{R}^{d_a}) \rightarrow \mathcal{U}(D; \mathbb{R}^{d_v})$, $\mathcal{Q}: \mathcal{U}(D; \mathbb{R}^{d_v}) \rightarrow \mathcal{U}(D; \mathbb{R}^{d_u})$, $d_v \geq d_a$, and $L$ is the number of layers.  

The layers $\mathcal{L}_l$ are defined as:  
\begin{equation*} \label{eq:five}  
\mathcal{L}_l(v)(x) = \sigma\left(W_l v(x) + (\mathcal{K}(a; \theta_l) v)(x)\right) \quad \forall x \in D,  
\end{equation*}  
where $W_l$ is a linear operator; $\sigma$ is a nonlinear activation function; and $\mathcal{K}(a; \theta_l)$ is a parameterized integral operator:  
\begin{equation*}\label{eq:six}  
(\mathcal{K}(a; \theta_l) v)(x) = \int_D \kappa_\theta(x, y; a(x), a(y)) v(y) \, dy \quad \forall x \in D.  
\end{equation*}  

The kernel $\kappa_\theta$ is a neural network with parameters $\theta\in\mathrm{\Theta}$ and can be defined in various ways, resulting in different types of neural operator architectures.

Note that the main difference between (\ref{eq:four}) and standard neural networks is that all operations are defined directly in function space (given that the activation functions, $\mathcal{P}$, and $\mathcal{Q}$ are interpreted through their extension to Nemytskii operators) and are therefore independent of the data discretization~[18, p. 10].

One of the most promising and well-studied methods for approximating PDE solutions is the family of architectures based on Fourier Neural Operators (FNO). The FNO architecture is schematically represented in Fig. 1 ([19]). 

\begin{figure}[h!]
    \centering
    \includegraphics[width=1\linewidth]{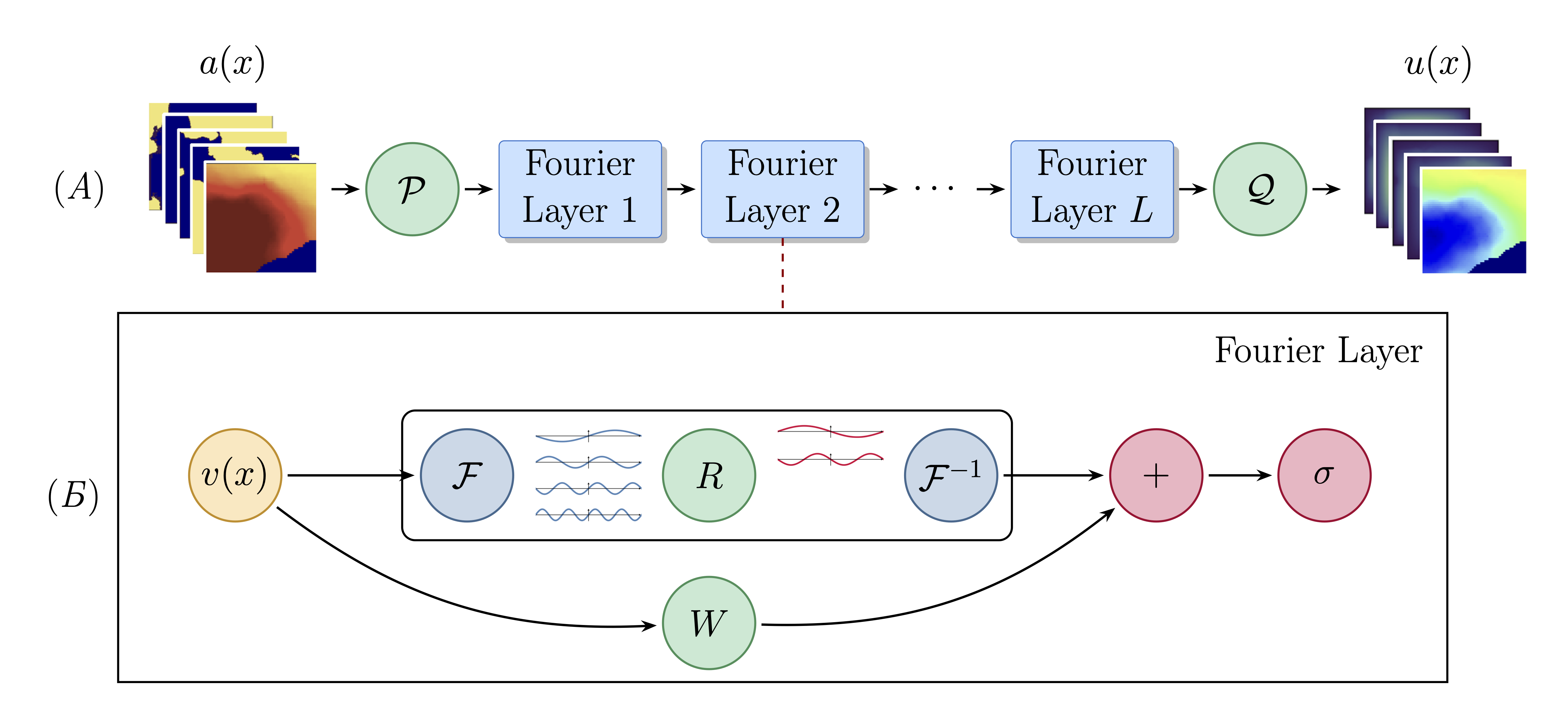}
    \vskip2mm {\small\emph{Fig. 1.}
 FNO architecture (\textit{A}) and Fourier layer (\textit{B}). \textit{A}: $a(x)$ is the input data, $\mathcal{P}$ and $\mathcal{Q}$ are fully-connected layers, $u(x)$ is the output data. \textit{B}: $\mathcal{F}$ and $\mathcal{F}^{-1}$ are the forward and inverse Fourier transforms, $R$ is the linear transform in the frequency domain, $W$ is a linear operator.}
    \label{fig:fno_layer}
\end{figure}

The kernel of the Fourier Neural Operator is expressed via convolution:
\begin{equation} \label{eq:seven}
    (\mathcal{K}(a;\theta_{l})v)(x)=\int_{D}\kappa _{\theta}(x-y)v(y)dy \ \forall x \in D,
\end{equation}
consequently, in the frequency domain, the kernel takes the form:
\begin{equation*} \label{eq:eight}
 (\mathcal{K}(\theta)v)(x)= \mathcal{F}^{-1} \bigg(R_{\theta}(k) \ \cdot \ \mathcal{F}(v)(k)\bigg)(x) \ \ \forall x \in D,
\end{equation*}
where \( \mathcal{F}\) and \( \mathcal{F}^{-1} \) are the forward and inverse Fourier transforms, respectively; $R_\theta\left(k\right)=\mathcal{F}\left(\kappa_\theta\right)\left(k\right)$ is the transformation coefficient tensor for $\kappa_\theta$; and \(k\) are the Fourier modes.

Note that in \eqref{eq:seven}, there is no explicit dependence of the kernel on $a$. Information about the input data is incorporated through the projection operator into the hidden space $\mathcal{P}$ and is passed between layers via the operator $W$.

Thus, the key feature of FNO is the parameterization of the integral operator's kernel in Fourier space, which allows for the effective capture of global dependencies in the data~[19]. Due to the use of the Fast Fourier Transform (FFT) algorithm, this method has logarithmic computational complexity, which has been confirmed, including for the approximation of parabolic PDEs~[24].

The architecture we have developed is based on Fourier Neural Operators and is an extension of the work in~[25, 26]. The Fourier transform within the method allows for the effective approximation of only a limited number of harmonics \(k\), which corresponds predominantly to the low-frequency component of the signal. In~[25], a method for hydrodynamic modeling of reservoir systems was implemented that compensates for this limitation by adding a convolutional network operator to the FNO architecture to account for local, high-frequency features of the data. When developing efficient neural operator algorithms that approximate nonlinear flow processes in reservoir systems, it is also necessary to consider the multidimensional spatio-temporal structure of the PDE solutions \eqref{eq:one}. One promising approach is the use of tensor methods for model parameter decomposition. Tensor decompositions can enhance model efficiency, as shown in~[27] for CNNs and in~[28] for neural operators. In~[26], a modified Fourier neural operator using factorized low-rank tensors (Tucker decomposition) was proposed. Such an architecture reduces the number of parameters, which allows for an increase in the number of Fourier modes to enhance the model's expressiveness and eliminates the need for adding finite-dimensional convolutional layers.

\subsection{Tensor Decompositions of the Neural Operator Parameters}
We will use Tensor-Train (TT) decompositions~[29] as a method for factorizing the parameter tensor. Unlike the Tucker decomposition, which requires \(\mathcal{O}(r^d)\) memory, the TT-decomposition is more efficient, as it requires \(\mathcal{O}(ndr^2)\) memory, where \(n\) is the tensor dimension, \(d\) is the number of dimensions, and \(r\) is the rank of the cores. This makes TT-decomposition preferable for high-dimensional problems.

For a neural network parameter tensor \({A} \in \mathbb{R}^{n_1 \times n_2 \times \dots \times n_d}\), the TT-decomposition is defined by a sequence of 3D tensors (cores) \(G_k \in \mathbb{R}^{r_{k-1} \times n_k \times r_k}\), where \(r_0 = r_d = 1\). Formally, this is expressed as:
\begin{equation}
\label{tt}
A(i_1, i_2, \dots, i_d) = \sum_{\alpha_1, \dots, \alpha_{d-1}} G_1(1, i_1, \alpha_1) \cdot G_2(\alpha_1, i_2, \alpha_2) \cdot \dots \cdot G_d(\alpha_{d-1}, i_d, 1),
\end{equation}
where \(\alpha_k\) are the summation indices, and \(r_k\) are the ranks of the cores.

TT-decomposition provides efficient compression and reconstruction of tensors while preserving their structure and significantly reducing the number of parameters. This property can improve the generalization ability of neural networks through structural regularization and thus reduce the risk of overfitting. Furthermore, TT-decomposition is particularly effective for working with spatio-temporal data, compactly representing complex dependencies in such structures.

\subsection{Controlling the Temporal Resolution of the Neural Operator}
Referring to the theory of dynamical systems~[30, 31], neural operators can be interpreted~[32] as specialized deep learning methods designed to approximate evolution operators acting in function spaces. Within the framework of Koopman operator theory~[31, 33], such operators allow for the construction of linear representations of nonlinear dynamics. In this approach, observable functions evolve under the action of the Koopman operator, which provides a global linearization (in the sense of replacing the original nonlinear system with a linear operator acting in the space of observable functions) of complex nonlinear systems. Fourier Neural Operators are represented in the frequency domain as linear transformations, which underscores their conceptual similarity to the elements of dynamical systems theory mentioned above. The authors of~[32] used Koopman operator theory to study the theoretical foundations of applying neural operators to model complex dynamical systems, including systems described by PDEs.

The initial data for training neural operators may not be available at the required temporal resolution (especially in the case of complex dynamical systems), which makes the task of approximating the evolution of the original processes more difficult. In such a case, it may be necessary to increase the model's internal temporal resolution~[34]. Let the input data be discretized in the time domain with a step \(\Delta t\). For the model to successfully approximate the solution operator of the target PDE while being trained on data with coarse temporal discretization, we define~\eqref{power_k} a composition of linear operators \(\mathcal{K}(\theta)\) in the frequency domain to introduce the \(r\)-th power of the Fourier integral operator:
\begin{equation}
\label{power_k}
(\mathcal{K}^r(\theta)v)(x) = \Bigl(\underbrace{\mathcal{K}(\theta) \circ \mathcal{K}(\theta) \circ \cdots \circ \mathcal{K}(\theta)}_{r \text{ times}} v\Bigr)(x)  \  \forall x \in D.
\end{equation}

In this case, the Fourier integral operator will take the form:
\begin{equation}
\label{power_k_2}
(\mathcal{K}^r(\theta) v)(x) = \Big(\mathcal{F}^{-1} \circ R_{\theta}^r(k) \circ \mathcal{F} (v)\Bigr)(x) \   \forall x \in D.
\end{equation}

With this formulation, modeling the evolution over a time interval \(\Delta t\) requires the operator \eqref{power_k_2} to perform \( r \) iterations (i.e., each iteration corresponds to a time interval of \(\Delta t/r\)). Thus, the introduction of the power \( r \) implies dividing the original discretization step \(\Delta t\) into \( r \) smaller steps \(\Delta t/r\), which can be interpreted~[34, p. 6] as increasing the temporal resolution of the trained neural operator. The scheme of the developed architecture is shown in Fig. 2.

\begin{figure}[h!]
    \centering
    \includegraphics[width=1\linewidth]{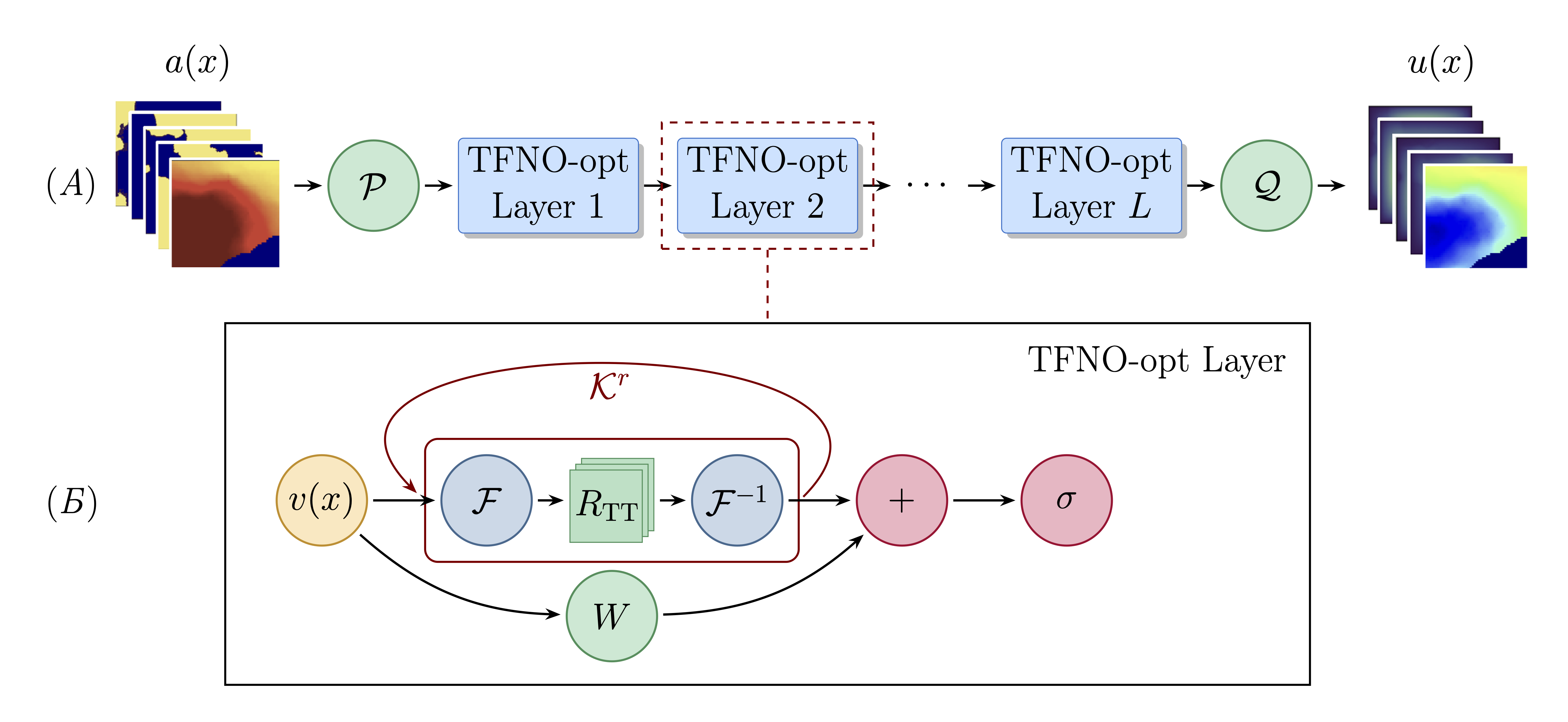}
    \vskip2mm {\small\emph{Fig. 2.}
 TFNO-opt architecture (\textit{A}) and TFNO-opt layer (\textit{B}). \textit{A}: $a(x)$ is the input data, $\mathcal{P}$ and $\mathcal{Q}$ are fully-connected layers, $u(x)$ is the output data. \textit{B}: $\mathcal{F}$ and $\mathcal{F}^{-1}$ are the forward and inverse Fourier transforms, $R_{TT}$ is the factorized linear operator, $W$ is a linear operator, $r$ is the power of the Fourier integral operator $\mathcal{K}$.}
    \label{fig:tfno_layer}
\end{figure}

\subsection{Development of a Loss Function for Neural Operator Training}
In this work, the relative Sobolev norm $H^1$ is used to develop the loss function for training the model: 

\begin{equation} 
\label{eq:eleven}
    L\left(\hat{y},y\right)= \Biggl( \sum\limits_{i=0}^k \frac{\left|\left|D^{i} \, \hat{y} -D^{i}  \, y\right|\right|_p^p}{\left|\left|D^{i} \, y \right|\right|_p^p} \Biggr)^{\frac{1}{p}}, \ k=1, \ p=2,
\end{equation}
where $\hat{y}$ is the predicted reservoir pressure; $y$ is the true reservoir pressure; $D^i$ is the differential operator of order $i$; and $p$ is the order of the norm. The derivatives are computed using the finite difference method.

The function \eqref{eq:eleven} has a normalization and regularization effect~[35], which is particularly important since reservoir pressure has different scales and can exhibit significant variations in space and time. Incorporating derivatives into the Sobolev norm allows for a more accurate consideration of the solution's smoothness, contributing to the correct reproduction of the physical nature of the modeled processes.

The final form of the loss function used is:
\begin{equation} 
\label{eq:loss_total}
\mathcal{L}(\theta) = \lambda \underbrace{L(\mathcal{G}_\theta(a), u)}_{\text{approximation}} \ + \ \gamma  \underbrace{L(\mathcal{Q} \circ \mathcal{P}(a), u_0)}_{\text{reconstruction}},
\end{equation}
where \(a\) is the input data; \(\mathcal{G}_\theta(a)\) is the model output; \(u\) is the true value; \(\mathcal{Q}\) and \(\mathcal{P}\) are the composition of the corresponding model layers applied to \(a\); \(u_0\) are the initial conditions (e.g., the pressure field at the initial time); \(\lambda, \gamma\in (0, \infty )\) are hyperparameters; and \(L\) is the function \eqref{eq:eleven}.

The separation of the contributions of approximation and reconstruction errors \eqref{eq:loss_total} has been investigated in~[30, 33]. The first term is responsible for approximation accuracy, while the second is responsible for preserving information about the initial conditions, which allows the model to reconstruct the initial data from the hidden space. This property plays a crucial role in correctly predicting system dynamics, as the reconstruction of initial conditions ensures accurate interpretation of input data by the model and helps maintain the physical consistency of the solution.

\section{Numerical Experiments}
The dataset used is based on the results of a three-dimensional hydrodynamic simulation of an underground gas storage (UGS) facility using a numerical simulator. Gas withdrawal seasons are used as the base periods for analysis. To increase the volume and representativeness of the training data, information related to injection seasons is also included. Thus, the entire sample includes historical data and results of numerical simulations of various gas withdrawal and injection scenarios with a time step of \(\Delta t=10\) days (decades). The UGS under consideration is a depleted gas field characterized by complex geometry, a porous reservoir structure, and a gas-drive regime. More than 100 wells are used for the operation of the UGS.

It is important to note that the training sample includes data for both gas withdrawal and injection seasons, whereas the validation and test samples contain only data for withdrawal seasons, which are of the greatest interest in this study, as these periods are characterized by increased demands for gas supply stability.

To evaluate the effectiveness of the proposed solutions, three neural operator architectures were developed and trained as part of the numerical experiments. The first model is a baseline Fourier Neural Operator. The second uses the Fourier integral operator developed in this work. The third is based on the second and has its parameter tensor factorized using TT-decomposition. All architectures include 4 Fourier layers and are limited to 16 frequency modes:

1) FNO (Fourier Neural Operator) – total parameters: 18.9 million;

2) FNO-opt – same as FNO, but with the modified Fourier integral operator~\eqref{power_k_2}. Power \(r=4\). Total parameters: 18.9 million;

3) TFNO-opt (proposed in this work) – same as FNO-opt but with a factorized parameter tensor \eqref{tt}, containing only about \(1\%\) of the total parameters – 0.2 million.

All models were trained using the loss function \eqref{eq:loss_total}. The training and validation results are presented in Fig. 3.
\begin{figure}[H]
    \centering
    \includegraphics[width=1\linewidth]{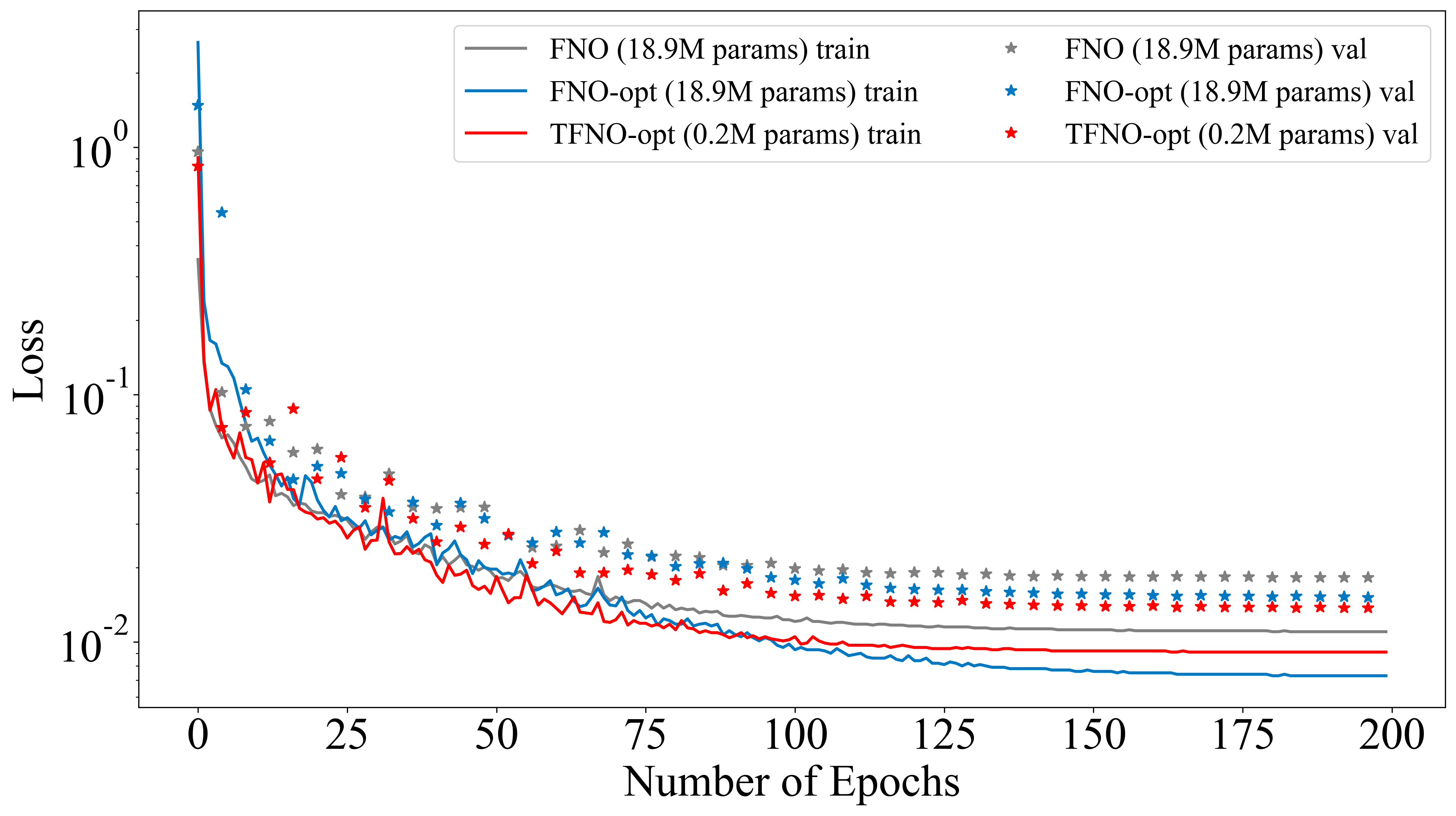}
    \vskip2mm {\small\emph{Fig. 3.}
Loss function values during the training and validation of the neural operators.}
    \label{fig:losses}
\end{figure}

Among the models listed above, the TFNO-opt architecture, which includes all the modifications developed in this work, showed the best performance while reducing the number of parameters by two orders of magnitude. It should be noted that the TT-decomposition of the model's parameters not only reduced their number but also improved the model's generalization ability, as evidenced by the loss function values on the validation set during training.

For an indirect assessment of the stability and training quality of the proposed model, the loss landscape~[36] was calculated (Fig. 4) on the test set, defined by the equation:
\begin{equation} 
\label{eq:loss_landscape}
f(\alpha, \beta) = L\big(\mathcal{G}_{\theta^* + \alpha \mathbf{d}_1 + \beta \mathbf{d}_2}(a), u\big),
\end{equation}
where \(L\) is the loss function; \(\theta^*\) is the weight tensor of the trained model; \(\mathbf{d}_1, \mathbf{d}_2 \in \mathbb{R}^n\) are two random normalized direction vectors; and \(\alpha, \beta \in \mathbb{R}\) are coefficients defining the displacement along the vectors.

\begin{figure}[H]
    \centering
    \includegraphics[width=1\linewidth]{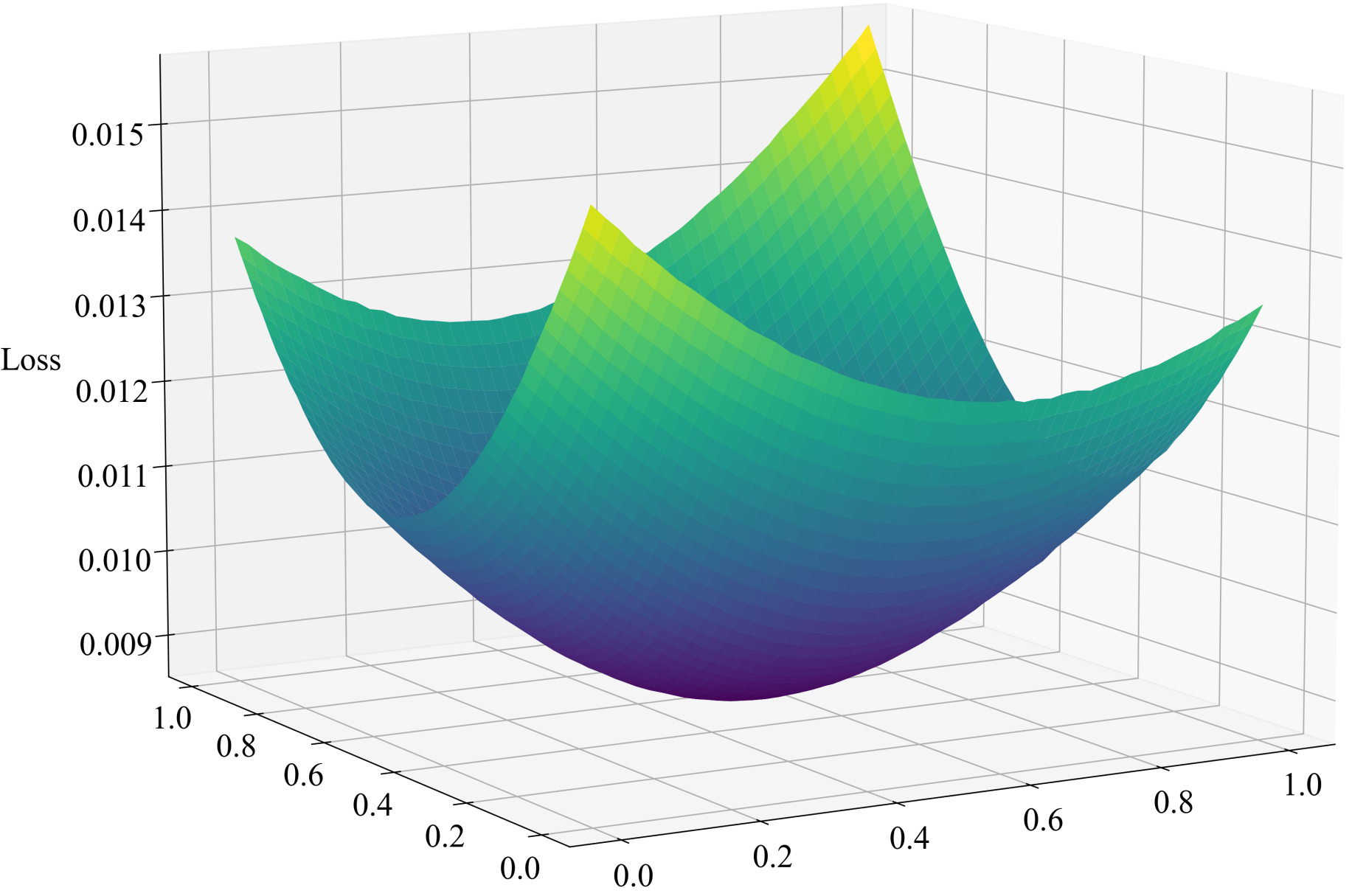}
    \vskip2mm {\small\emph{Fig. 4.}
Loss landscape of the trained model (TFNO-opt).}
    \label{fig:loss_landscape}
\end{figure}

The loss landscape \eqref{eq:loss_landscape}, constructed on the test set, is visually characterized by a smooth, convex structure with a distinct minimum. Such a shape may indicate that the model exhibits a degree of stability to perturbations, which is critically important for optimal control problems.

Fig. 5 shows a scatter plot of the normalized (scaled to the range of 0 to 1) reservoir pressure obtained using the proposed model, compared with the results of numerical simulation on the test data.

\begin{figure}[h!]
    \centering
    \includegraphics[width=1\linewidth]{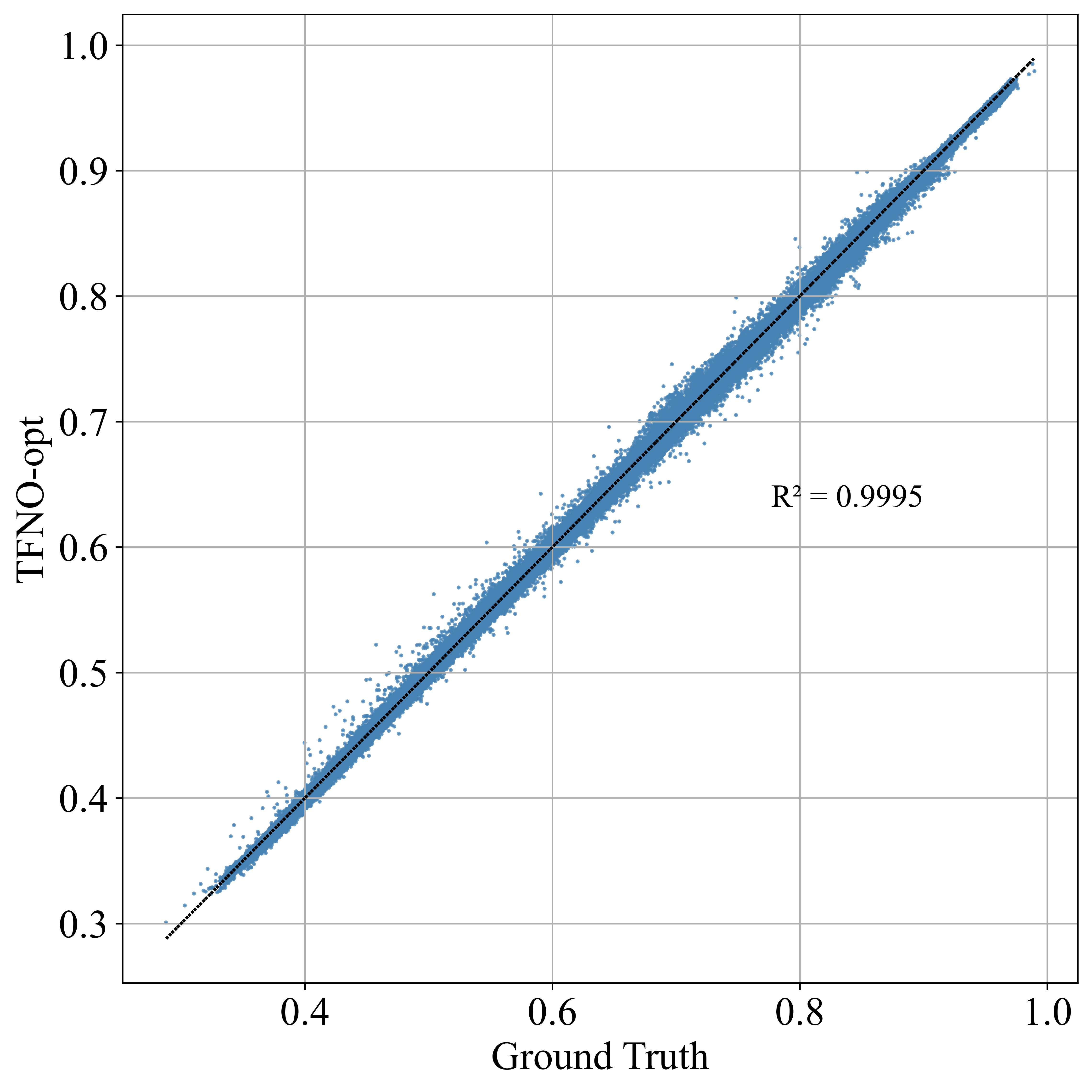}
    \vskip2mm {\small\emph{Fig. 5.}
    Scatter plot of normalized reservoir pressure.}
    \label{fig:loss_landscape_scatter}
\end{figure}
This plot shows that the distribution generated by the neural operator on the test data for each point in the reservoir is very close to the distribution obtained using the hydrodynamic model. The coefficient of determination is $R^2 = 0.9995$.

It should be noted that the time required to compute a single scenario is a fraction of a second, which is at least six orders of magnitude faster than a similar calculation on a numerical simulator.

A comparison of the reservoir pressure field dynamics, as predicted by the neural operator, with the results from the numerical simulation (on test data) is presented in Fig. 6. The time step indicates the ordinal number of the ten-day period (decade) within the gas withdrawal season. A fragment of the UGS is shown (in the $i,j$ plane).

The calculations of the trained model (TFNO-opt) reproduce both global and local features of the reservoir pressure distribution. The values of the absolute error in modeling the reservoir pressure on the test set are insignificant. Thus, the results show high consistency between the neural operator calculations and the numerical simulation results, confirming the effectiveness of the proposed architecture for modeling reservoir pressure dynamics.

\begin{figure}[H]
    \centering
    \includegraphics[width=1.02\linewidth]{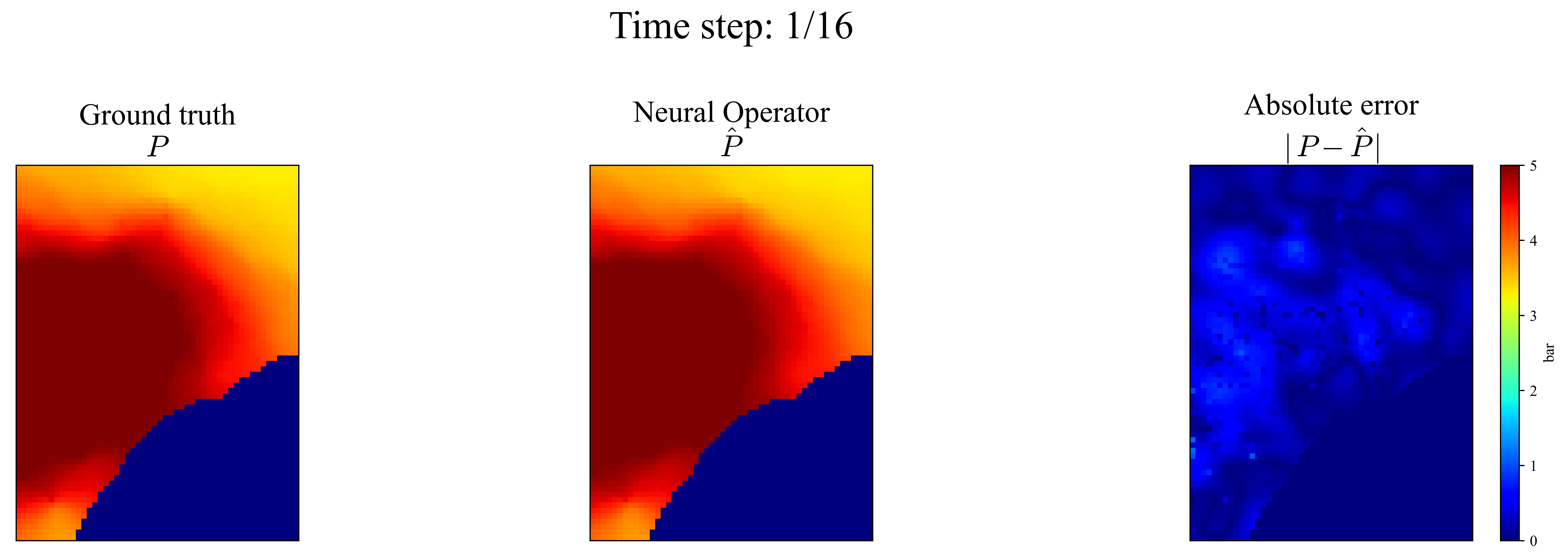}
\end{figure}
\begin{figure}[H]
    \centering
    \includegraphics[width=1.02\linewidth]{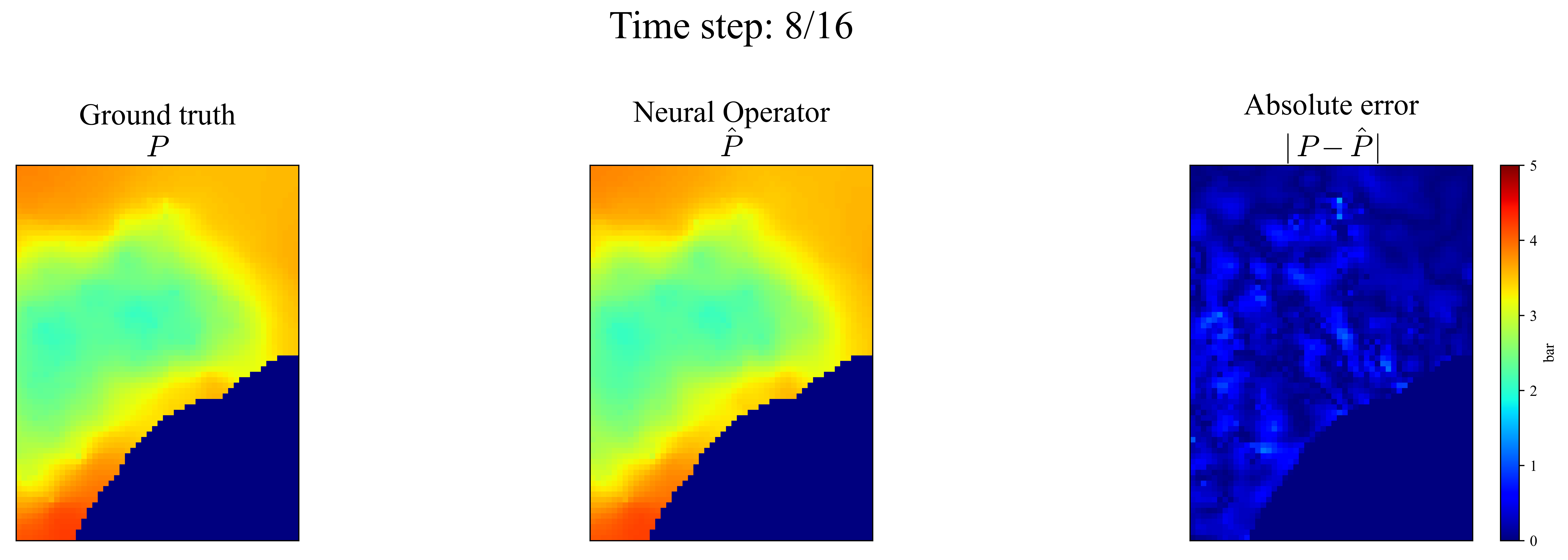}
\end{figure}
\begin{figure}[H]
    \centering
    \includegraphics[width=1.02\linewidth]{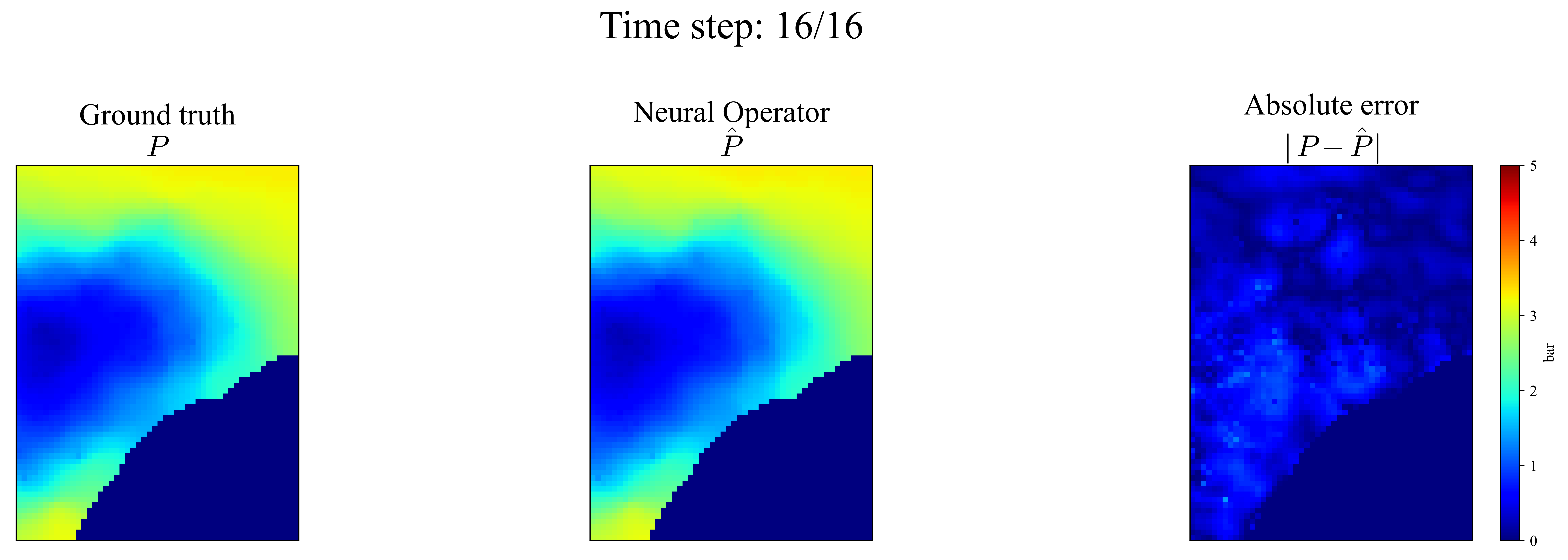}
    \vskip2mm {\small\emph{Fig. 6.}
     Comparison of the reservoir pressure field from the numerical model, the neural operator (TFNO-opt), and the absolute error on the test set.}
    \label{fig:field_comparison}
\end{figure}

\section{Conclusion}
A neural operator architecture (TFNO-opt) adapted for modeling transient flow in reservoir systems was developed. 

The \textit{efficiency and accuracy of the modeling are improved} relative to the baseline FNO architecture through: 1) the adjustable temporal resolution of the integral operator's kernel; 2) the tensor decomposition of the model's parameters; and 3) the application of the Sobolev \(H^1\) norm as the loss function.

The proposed architecture provides a computational speed-up of six orders of magnitude relative to numerical reservoir models, as confirmed by computational experiments on the UGS modeling problem. This opens up possibilities for the \textit{optimization} of operational regimes and the \textit{control} of reservoir systems.

Promising directions for future research include:

1) Developing a control algorithm and analyzing the efficiency of TFNO-opt in problems of optimizing the operational regimes of reservoir systems;

2) Integrating physical constraints (e.g., material balance) to guarantee the observance of conservation laws, even when modeling extreme scenarios;

3) Investigating the possibility of developing universal ("foundational") models that encapsulate the underlying physics of flow processes, with subsequent adaptation to specific assets. Such an approach could eliminate the need to create comprehensive datasets for each asset, requiring only a small amount of data for model fine-tuning.

\vskip6mm
\section*{References}

\vskip3mm
{\footnotesize

1. Aziz K., Settari A. \emph{Petroleum reservoir simulation}. London, New York, Applied Science Publ., 1979, 497 p.

2. LeCun Y., Bengio Y., Hinton G. Deep learning. \emph{Nature}, 2015, vol. 521, no. 7553, pp. 436–444. https://doi.org/10.1038/nature14539

3. Goodfellow I., Bengio Y., Courville A. \emph{Deep learning}. Cambridge, Massachusetts, The MIT Press, 2016, 775 p.

4. Cybenko G. Approximation by superpositions of a sigmoidal function. \emph{Mathematics of Control, Signals, and Systems}, 1989, vol. 2, no. 4, pp. 303–314. 

\noindent https://doi.org/10.1007/BF02551274

5. Hornik K., Stinchcombe M., White H. Multilayer feedforward networks are universal approximators. \emph{Neural Networks}, 1989, vol. 2, no. 5. pp. 359–366. https://doi.org/10.1016/0893-6080(89)90020-8

6. Chen T., Chen H. Approximations of continuous functionals by neural networks with application to dynamic systems. \emph{IEEE Transactions on Neural Networks}, 1993, vol. 4, no. 6, pp. 910–918.

\noindent https://doi.org/10.1109/72.286886

7. Chen T., Chen H. Universal approximation to nonlinear operators by neural networks with arbitrary activation functions and its application to dynamical systems. \emph{IEEE Transactions on Neural Networks}, 1995, vol. 6, no 4, pp. 911–917. https://doi.org/10.1109/72.392253

8. Yarotsky D. Error bounds for approximations with deep ReLU networks. \emph{Neural Networks}, 2017, vol. 94, pp. 103–114. https://doi.org/10.1016/j.neunet.2017.07.002

9. Han J., Jentzen A., E W. Solving high-dimensional partial differential equations using deep learning. \emph{Proceedings of the National Academy of Sciences}, 2018, vol. 115, no 34. pp. 8505–8510. 

\noindent https://doi.org/10.1073/pnas.1718942115

10. Jakubovitz D., Giryes R., Rodrigues M. R. D. Generalization error in deep learning. \emph{Compressed Sensing and its Applications}, Cham, Springer International Publishing, 2019, pp. 153–193. 

\noindent https://doi.org/10.1007/978-3-319-73074-5\textunderscore5

11. Guo X., Li W., Iorio F. Convolutional neural networks for steady flow approximation. \emph{Proceedings of the 22\textsuperscript{nd} ACM SIGKDD International Conference on Knowledge Discovery and Data Mining}. San Francisco, California, USA, ACM Publ., 2016, pp. 481–490. https://doi.org/10.1145/2939672.2939738

12. Tang M., Liu Y., Durlofsky L. J. A deep-learning-based surrogate model for data assimilation in dynamic subsurface flow problems. \emph{Journal of Computational Physics}, 2020, vol. 413, art. no. 109456. https://doi.org/10.1016/j.jcp.2020.109456

13. Zhong Z., Sun A. Y., Jeong H. Predicting CO\textsubscript{2} plume migration in heterogeneous formations using conditional deep convolutional generative adversarial network. \emph{Water Resources Research}, 2019, vol. 55, no. 7, pp. 5830–5851. https://doi.org/10.1029/2018WR024592

14. Berg J., Nyström K. A unified deep artificial neural network approach to partial differential equations in complex geometries. \emph{Neurocomputing}, 2018, vol. 317, pp. 28–41. 

\noindent https://doi.org/10.1016/j.neucom.2018.06.056

15. Raissi M., Perdikaris P., Karniadakis G. E. Physics-informed neural networks: A deep learning framework for solving forward and inverse problems involving nonlinear partial differential equations. \emph{Journal of Computational Physics}, 2019, vol. 378, pp. 686–707. https://doi.org/10.1016/j.jcp.2018.10.045

16. Baydin A. G., Pearlmutter B. A., Radul A. A., Siskind J. M. Automatic differentiation in machine learning: a survey. \emph{Journal of Machine Learning Research}, 2018, vol. 18, pp. 1–43. 

\noindent https://doi.org/10.48550/ARXIV.1502.05767

17. Lu L., Jin P., Pang G., Zhang Z., Karniadakis G. E. Learning nonlinear operators via DeepONet based on the universal approximation theorem of operators. \emph{Nature Machine Intelligence}, 2021, vol. 3, no. 3, pp. 218–229. https://doi.org/10.1038/s42256-021-00302-5

18. Kovachki N., Li Z., Liu B., Azizzadenesheli K., Bhattacharya K., Stuart A., Anandkumar A. Neural operator: Learning maps between function spaces with applications to PDEs. \emph{The Journal of Machine Learning Research}, 2023, vol. 24, no. 89. pp. 1–97. 
\noindent https://doi.org/10.48550/ARXIV.2108.08481

19. Li Z., Kovachki N., Azizzadenesheli K., Liu B., Bhattacharya K., Stuart A., Anandkumar A. Fourier Neural Operator for parametric partial differential equations. \emph{arXiv preprint. arXiv: 2010.08895},  2020, https://doi.org/10.48550/ARXIV.2010.08895

20. Ole{\u{\i}}nik O.~A., Kalashnikov A.~S., Zhou Yu-lin. Zadacha Koshi i kraevye zadachi dlia uravnenii tipa nestatsionarnoi fil'tratsii [The Cauchy problem and boundary problems for equations of the type of non-stationary filtration]. \emph{Proceedings of the Academy of Sciences of the USSR. Series Mathematics}, 1958, vol. 22, no. 5, pp. 667–704. (In Russian).

21. Vazquez J. L. \emph{The porous medium equation: Mathematical theory}. 1\textsuperscript{st} ed. Oxford, Oxford University Press, 2006. 561 p. https://doi.org/10.1093/acprof:oso/9780198569039.001.0001

22. Ertekin T., Abou-Kassem J. H., King G. R. \emph{Basic applied reservoir simulation}. Richardson, Tex, Society of Petroleum Engineers Publ., 2001, 406 p.

23. Chen Z. \emph{Reservoir simulation: Mathematical techniques in oil recovery}. Society for Industrial and Applied Mathematics Publ., 2007, 248 p. https://doi.org/10.1137/1.9780898717075

24. Lanthaler S., Molinaro R., Hadorn P., Mishra S. Nonlinear reconstruction for operator learning of PDEs with discontinuities. \emph{arXiv preprint. arXiv: 2210.01074}, 2022, 

\noindent https://doi.org/10.48550/ARXIV.2210.01074

25. Sirota D. D., Gushchin K.A., Khan S.A., Kostikov S.L., Butov K.A. Neural operators for hydrodynamic modeling of underground gas storage facilities. \emph{Russian Technological Journal}, 2024, vol. 12, no. 6, pp. 102–112. https://doi.org/10.32362/2500-316X-2024-12-6-102-112

26. Sirota D. D., Gushchin K.A., Khan S.A., Kostikov S.L., Butov K.A. Neural operators for hydrodynamic modeling of underground gas storages. \emph{Moscow University Physics Bulletin}, 2024, vol. 79, no. S2, pp. S922–S934. 
https://doi.org/10.3103/S0027134924702382

27. Kossaifi J., Toisoul A., Bulat A., Panagakis Y., Hospedales T., Pantic M. Factorized higher-order CNNs with an application to spatio-temporal emotion estimation. \emph{2020 IEEE/CVF Conference on Computer Vision and Pattern Recognition (CVPR)}. Seattle, WA, USA, IEEE Publ., 2020, pp. 6059–6068. https://doi.org/10.1109/CVPR42600.2020.00610

28. Kossaifi J., Kovachki N., Azizzadenesheli K., Anandkumar A. Multi-grid tensorized fourier neural operator for high-resolution PDEs. \textit{arXiv preprint. arXiv:2310.00120}, 2023.

\noindent https://doi.org/10.48550/arXiv.2310.00120.

29. Oseledets I. V. Tensor-train decomposition. \emph{SIAM Journal on Scientific Computing}, 2011, vol. 33, no. 5, pp. 2295–2317. https://doi.org/10.1137/090752286

30. Brunton S. L., Kutz J. N. \emph{Data-Driven Science and Engineering: Machine Learning, Dynamical Systems, and Control}, 2\textsuperscript{nd} ed. Cambridge, Cambridge University Press, 2022, 00p. 

\noindent https://doi.org/10.1017/9781009089517

31. Nakao H., Mezić I. Spectral analysis of the Koopman operator for partial differential equations. \emph{Chaos: An Interdisciplinary Journal of Nonlinear Science}, 2020, vol. 30, no. 11, art. no. 113131.

\noindent https://doi.org/10.1063/5.0011470

32. Xiong W., Ma M., Huang X., Zhang Z., Sun P., Tian Y. KoopmanLab: Machine learning for solving complex physics equations. \emph{APL Machine Learning}, 2023, vol. 1, no. 3, art. no. 036110. https://doi.org/10.1063/5.0157763

33. \emph{The Koopman Operator in Systems and Control: Concepts, Methodologies, and Applications}. Eds: Mauroy A., Mezić I., Susuki Y. Cham, Springer International Publishing, 2020, vol. 484, 00p. 

\noindent https://doi.org/10.1007/978-3-030-35713-9

34. Xiong W., Huang X., Zhang Z., Deng R.,  Sun P., Tian Y. Koopman neural operator as a mesh-free solver of non-linear partial differential equations. \emph{arXiv preprint. arXiv:2301.10022}, 2024. 

\noindent https://doi.org/10.48550/arXiv.2301.10022

35. Czarnecki W. M., Osindero S., Jaderberg M., Świrszcz G., Pascanu R. Sobolev training for neural networks. \emph{arXiv preprint. arXiv:1706.04859}, 2017. 

\noindent https://doi.org/10.48550/arXiv.1706.04859

36. Li H., Xu Z., Taylor G., Studer C., Goldstein T. Visualizing the loss landscape of neural nets. \emph{arXiv preprint. arXiv:1712.09913}, 2018. https://doi.org/10.48550/arXiv.1712.09913

\end{document}